\documentclass[sigconf]{acmart}
\pdfoutput=1
\def\BibTeX{{\rm B\kern-.05em{\sc i\kern-.025em b}\kern-.08emT\kern-.1667em\lower.7ex\hbox{E}\kern-.125emX}}

\usepackage{booktabs} 
\usepackage{multirow}

\usepackage{graphicx}
\usepackage{caption}
\usepackage{subcaption}

\usepackage{algorithm,algorithmicx,algpseudocode}
\usepackage{tikz,pgfplots}
\usetikzlibrary{arrows,automata,positioning,decorations.pathreplacing}
\usepackage{booktabs,subcaption,amsfonts,dcolumn}
\usetikzlibrary{patterns}

\usepackage{balance}

\setcopyright{rightsretained}

\acmDOI{10.475/123_4}

\acmISBN{123-4567-24-567/08/06}

\renewcommand\footnotetextcopyrightpermission[1]{} 
\acmArticle{4}
\acmPrice{15.00}

\begin{document}
\title{Graph Attention Auto-Encoders}

\author{Amin Salehi}
\affiliation{%
  \institution{Computer Science and Engineering \\ Arizona State University}
  \city{Tempe}
  \state{Arizona}
}
\email{asalehi1@asu.edu}

\author{Hasan Davulcu}
\affiliation{%
    \institution{Computer Science and Engineering \\ Arizona State University}
    \city{Tempe}
    \state{Arizona}
}
\email{hdavulcu@asu.edu}

\begin{abstract}
    
Auto-encoders have emerged as a successful framework for unsupervised learning. However, conventional auto-encoders are incapable of utilizing explicit relations in structured data. To take advantage of relations in graph-structured data, several graph auto-encoders have recently been proposed, but they neglect to reconstruct either the graph structure or node attributes. In this paper, we present the graph attention auto-encoder (GATE), a neural network architecture for unsupervised representation learning on graph-structured data. Our architecture is able to reconstruct graph-structured inputs, including both node attributes and the graph structure, through stacked encoder/decoder layers equipped with self-attention mechanisms. In the encoder, by considering node attributes as initial node representations, each layer generates new representations of nodes by attending over their neighbors' representations. In the decoder, we attempt to reverse the encoding process to reconstruct node attributes. Moreover, node representations are regularized to reconstruct the graph structure. Our proposed architecture does not need to know the graph structure upfront, and thus it can be applied to inductive learning. Our experiments demonstrate competitive performance on several node classification benchmark datasets for transductive and inductive tasks, even exceeding the performance of supervised learning baselines in most cases.

\end{abstract}

%
%
%
%
%

\maketitle

\section{Introduction}
Low-dimensional vector representations of nodes in graphs have demonstrated their utility in a broad range of machine learning tasks. Such tasks include node classification \cite{grover2016node2vec}, recommender systems \cite{ying2018graph}, community detection \cite{wang2017community}, graph visualization \cite{perozzi2014deepwalk, tang2015line}, link prediction \cite{wei2017cross} and relational modeling \cite{schlichtkrull2018modeling}. Accordingly, there has been a surge of research to learn better node representations. However, most of the proposed methods \cite{grover2016node2vec,
    belkin2002laplacian, he2004locality, ahmed2013distributed, cao2015grarep, ou2016asymmetric, perozzi2014deepwalk, grover2016node2vec, perozzi2016walklets, chamberlain2017neural, tian2014learning, wang2016structural, tang2015line, cao2016deep, chen2018harp} only utilize the graph structure while nodes in real-world graphs usually come with a rich set of attributes (i.e. features). Typical examples are users in social networks, scientific articles in citation networks,
protein molecules in biological networks and web pages on the Internet.  

Significant efforts have been made \cite{huang2017label, yang2016revisiting, defferrard2016convolutional, monti2017geometric, kipf2017semi, velickovic2018graph, hamilton2017inductive} to utilize node attributes for graph representation learning. Nevertheless, the most successful methods, notably graph convolutional networks \cite{kipf2017semi} and graph attention networks \cite{velickovic2018graph}, depend on label information, which is not available in many real-world applications. Moreover, the process of annotating data suffers from many limitations, such as annotators’ subjectivity, reproducibility, and consistency.

To avoid the challenges of annotating data, several unsupervised graph embedding methods \cite{kipf2016variational, duran2017learning, pan2018adversarially, velivckovic2018deep, gao2018deep, huang2017accelerated, yang2015network} have been proposed, but these methods suffer from at least one of the three following problems. First, despite utilizing node features, some of these models \cite{kipf2016variational, pan2018adversarially, gao2018deep} heavily depend on the graph structure. This hinders their capability to fully exploit node features. Second, many \cite{gao2018deep, huang2017accelerated, yang2015network} are not capable of inductive learning, which is crucial to encounter unseen nodes (e.g., new users in social networks, recently published scientific articles and new web pages on the Internet). Third, even though some efforts have been made \cite{velivckovic2018deep, hamilton2017inductive} to address inductive learning tasks, they are not unified architectures for both transductive and inductive tasks.


Auto-encoders have recently become popular for unsupervised learning due to their ability to capture complex relationships between input's attributes through stacked non-linear layers \cite{baldi2012autoencoders, bengio2013representation}. However, conventional auto-encoders are not able to take advantage of explicit relations in structured data. To utilize relations in graph-structured data, several graph auto-encoders \cite{kipf2016variational, wang2017mgae, pan2018adversarially} have been proposed. Although the encoders in these models fully utilize graph-structured inputs, the decoders neglect to reconstruct either the graph structure or node attributes.


Another successful neural network paradigm is the attention mechanism \cite{bahdanau2014neural}, which has been extremely useful in tackling many machine learning tasks \cite{chorowski2015attention, chen2016attention, wang2018non}, particularly sequence-based tasks \cite{vaswani2017attention, luong2015effective, dehghani2018universal, rush2015neural}. The state-of-the-art attention mechanism is self-attention or intra-attention, which computes the representation of an input (e.g., a set or sequence) by focusing on its most relevant parts. Self-attention has been successfully applied to a variety of tasks including machine translation \cite{vaswani2017attention}, video classification \cite{wang2018non} and question answering \cite{dehghani2018universal}. Nonetheless, the majority of these efforts target supervised learning tasks, and few efforts \cite{devlin2018bert, he2017unsupervised} are made to tackle unsupervised learning tasks. In graph representation learning, to our knowledge, the only proposed attention-based method uses supervised learning  \cite{velickovic2018graph}.


In this work, we present a novel graph auto-encoder to learn node representations within graph-structured data (i.e., attributed graphs) in an \textit{unsupervised manner}. Our auto-encoder takes in and reconstructs node features by utilizing the graph structure through stacked encoder/decoder layers. In the encoder, node attributes are fed into stacked layers to generate node representations. By considering node features as initial node representations, each encoder layer generates new representations of nodes by utilizing neighbors' representations according to their relevance, which is determined by a graph attention mechanism. In the decoder, we aim to reverse the entire encoding process to reconstruct node attributes. To this end, each decoder layer attempts to reverse the process of its corresponding encoder layer. Moreover, node representations are regularized to reconstruct the graph structure. To our knowledge, no auto-encoder is capable of reconstructing both node attributes and the graph structure. Our architecture can also be applied to inductive learning tasks since it doesn't need to know the graph structure upfront.

Our key contributions are summarized as follows:

\begin{itemize}
    \item We propose a novel graph auto-encoder for unsupervised representation learning on graph-structured data by reconstructing both node features and the graph structure.
    \item We utilize self-attention for unsupervised attributed graph representation learning.
    \item We present a unified neural architecture capable of both transductive and inductive learning.
\end{itemize}

The rest of the paper is organized as follows. We review related work in Section \ref{sec:relted_work}. In Section \ref{sec:problem_statement}, we formally define the problem of unsupervised representation learning on graph-structured data. Section \ref{sec:architecture} presents the architecture of our proposed graph auto-encoder. In Section \ref{sec:evaluation}, we quantitatively and qualitatively evaluate GATE using several benchmark datasets for both transductive and inductive learning tasks. Section \ref{sec:conclusions} concludes the paper.

\section{Related Work}
\label{sec:relted_work}

\subsection{Graph Representation Learning}
Most of the graph embedding methods fall into one of the following three categories: factorization based, random walk based, and auto-encoder based approaches. 

Factorization based approaches are inspired by matrix factorization methods, which assume that the data lies in a low dimensional manifold. Laplacian Eigenmaps \cite{belkin2002laplacian} and LPP \cite{he2004locality} rely on eigendecomposition to preserve the local manifold structure. Due to expensive eigendecomposition operations, these methods face difficulty to tackle large-scale graphs. To alleviate this problem, several techniques---notably the Graph Factorization (GF) \cite{ahmed2013distributed}, GraRep \cite{cao2015grarep} and HOPE \cite{ou2016asymmetric}---have been proposed. These methods differ mainly in their node similarity calculation. The graph factorization computes node similarity based on the first-order proximities directly extracted from the adjacency matrix. To capture more accurate node similarity, GraRep and HOPE utilize the high-order proximities obtained from different powers of the adjacency matrix and similarity measures (i.e., cosine similarity)  respectively.

Random walk based approaches assume a pair of nodes to be similar if they are close in simulated random walks over the graph. Therefore, node similarity is stochastically computed in contrast to the deterministic approach used by factorization based methods. DeepWalk \cite{perozzi2014deepwalk} and node2vec \cite{grover2016node2vec} are the most successful methods in this category and differ primarily in their random walk generation. DeepWalk simulates uniform random walks while node2vec relies on a biased random walk generation. Preozzi et al. \cite{perozzi2016walklets} extend DeepWalk to encode multiscale node relationships in the graph. In contrast to DeepWalk and node2vec, which embed nodes in the Euclidean space, Chamberlan et al. \cite{chamberlain2017neural} utilize the hyperbolic space.

Factorization based and random walk based approaches adopt shallow models, which are incapable of capturing complex graph structures. To solve this problem, auto-encoder based approaches are proposed to capture non-linear graph structures by using deep neural networks. Tian et al. \cite{tian2014learning} present a stacked sparse auto-encoder to embed nodes by reconstructing the adjacency matrix. Moreover, Wang et al. \cite{wang2016structural} propose a stacked auto-encoder, which reconstructs the second-order proximities by using the first-order proximities as a regularization. Cao et al. \cite{cao2016deep} use stacked denoising auto-encoder to reconstruct the pointwise mutual information matrix.

Although the majority of graph embedding methods fall into one of the aforementioned categories, there are still some exceptions. For instance, LINE \cite{tang2015line} is a successful shallow embedding method to preserve both the local and global graph structures. Another example is HARP \cite{chen2018harp}, which introduces a graph processing step coarsening a graph into smaller graphs at different levels of granularity, and then it embeds them from the smaller graph to the largest one (i.e., the original graph) by using one of the graph embedding methods (i.e., DeepWalk, node2vec, and LINE). 

\subsection{Attributed Graph Representation Learning}
The graph embedding methods described in the previous section only utilize the graph structure to learn node representations. However, nodes in real-world graphs usually come with a rich set of attributes. To take advantage of node features, many attributed graph embedding methods have been proposed, which fall into two main categories: supervised and unsupervised approaches.

Supervised attributed graph embedding approaches embed nodes by utilizing label information. For example, Huang et al. \cite{huang2017label} propose a supervised method leveraging spectral techniques to project the adjacency matrix, node feature matrix, and node label matrix into a common vector space.  Hamilton et al. \cite{hamilton2017inductive} present four variants of GraphSAGE, a framework to compute node embeddings in an inductive manner. Many approaches address graphs with partial label information. For example, Graph Convolution Network (GCN) \cite{kipf2017semi} incorporates spectral convolutions into neural networks. Graph Attention Network (GAT) \cite{velickovic2018graph} utilizes an attention mechanism to determine the influence of neighboring nodes in final node representations.

The unsupervised attributed graph embedding methods address the lack of label information, which exists in many real-world applications. Yang et al. \cite{yang2015network} and Huang et al. \cite{huang2017accelerated} propose matrix factorization methods to combine the graph structure and node attributes. Moreover, Kipf et al. \cite{kipf2016variational} propose two graph auto-encoders utilizing graph convolution networks. Pan et al. \cite{pan2018adversarially} also introduce a graph-encoder based on an adversarial approach. For graph clustering, Wang et al. \cite{wang2017mgae} present a graph auto-encoder, which is able to reconstruct node features. However, these auto-encoders reconstruct either the graph structure or node attributes instead of both. To alleviate this limitation, Gao et al. \cite{gao2018deep} propose a framework consisting of two conventional auto-encoders, which reconstruct the graph structure and node attributes separately. These two auto-encoders are regularized in a way that their learned representations of neighboring nodes are similar. However, their framework does not fully leverage the graph structure due to the incapability of conventional auto-encoders in utilizing explicit relations in structured data. Most of the aforementioned unsupervised methods are not designed for inductive learning, which is crucial to encounter unseen nodes. Velivckovic et al. \cite{velivckovic2018deep} and Hamilton et al. \cite{hamilton2017inductive} propose unsupervised models for tackling inductive tasks, but their models are not unified frameworks for both transductive and inductive tasks.



%
%

\section{Problem Statement}
\label{sec:problem_statement}
In this section, we present the notations used in the paper and formally define the problem of unsupervised node representation learning on graph-structured data. We use bold upper-case letters for matrices (e.g., $\mathbf{X}$), bold lowercase letters for vectors (e.g., $\mathbf{x}$), and calligraphic fonts for sets (e.g., $\mathcal{N}$). Moreover, we represent the transpose of a matrix $\mathbf{X}$ as $\mathbf{X}^T$. The $i^\text{th}$ element of vector $\mathbf{x}$ is denoted by $\mathbf{x}_i$. $\mathbf{X}_{ij}$ denotes the entry of matrix $\mathbf{X}$ at the $i^\text{th}$ row and the $j^\text{th}$ column. Table \ref{tab:notation} summarizes the main notations used in the paper.

In the attributed graph representation learning setup, we are provided with the node feature matrix $\mathbf{X} = [\mathbf{x}_1, \mathbf{x}_2,..., \mathbf{x}_N]$, where $N$ is the number of nodes in the graph and $\mathbf{x}_i \in \mathbb{R}^F$ corresponds to the $i^\text{th}$ column of matrix $\mathbf{X}$, denoting the features of node $i$. We are also given the adjacency matrix $\mathbf{A} \in \mathbb{R}^{N \times N}$, representing the relations between nodes. Even though the matrix $\mathbf{A}$ may consist of real numbers, in our experiments, we assume the graph is unweighted and includes self-loops, i.e., $\mathbf{A}_{ij}=1$ if there is an edge between node $i$ and node $j$ in the graph or $i$ equals $j$, and $\mathbf{A}_{ij}=0$ otherwise.
 
Given the node feature matrix $\mathbf{X}$ and the adjacency matrix $\mathbf{A}$, our objective is to learn node representations in the form of matrix $\mathbf{H} = [\mathbf{h}_1, \mathbf{h}_2,..., \mathbf{h}_N]$, where $\mathbf{h}_i \in \mathbb{R}^D$ corresponds to the $i^\text{th}$ column of matrix $\mathbf{H}$, denoting the representation of node $i$.

\section{Architecture}
\label{sec:architecture}
In this section, we illustrate the architecture of the graph attention auto-encoder. First, we present the encoder and decoder to show how our auto-encoder reconstructs node features using the graph structure. Then, we describe the proposed loss function, which learns node representations by minimizing the reconstruction loss of node features and the graph structure. In the end, we present the matrix formulation of GATE, as well as its time and space complexities.

\begin{table}
    \centering
    \caption{The main notations used in the paper.}
    \label{tab:notation}
    \small
    \begin{tabular}{|l|p{5.7cm}|} \hline
        \textbf{Notations} & \textbf{Definitions}\\ \hline \hline
        
        $N$ & The number of nodes in the graph \\
        $E$ & The number of edges in the graph \\
        $L$ & The number of layers \\
        $d^{(k)}$ & The number of node representation dimensions in the $k^\text{th}$ encoder/decoder layer  \\ 
        $F$ & The number of node features ($d^{(0)} = F$) \\
        
        $P$ & The number of iterations (i.e., epochs) \\
        
        $\mathbf{A} \in \mathbb{R}^{N \times N}$ & The adjacency matrix \\
        
        $\mathbf{H}^{(k)} \in \mathbb{R}^{d^{(k)} \times N }$ & The node representation matrix generated by the $k^\text{th}$ encoder layer \\    
        
        $\widehat{\mathbf{H}}^{(k)} \in \mathbb{R}^{d^{(k)} \times N }$ & The node representation matrix reconstructed by the $k^\text{th}$ decoder layer \\    
        
        $\mathbf{H} \in \mathbb{R}^{d^{(L)} \times N}$ & The node representation matrix ($\mathbf{H} = \mathbf{H}^{(L)} = \widehat{\mathbf{H}}^{(L)}$)   \\
        
        $\mathbf{X} \in \mathbb{R}^{F \times N}$ & The node feature matrix ($\mathbf{H}^{(0)}=\mathbf{X}$) \\
        
        $\widehat{\mathbf{X}} \in \mathbb{R}^{F \times N}$ & The reconstructed node feature matrix ( $\widehat{\mathbf{X}} = \widehat{\mathbf{H}}^{(0)}$) \\ 
        
        $\mathbf{C}^{(k)} \in \mathbb{R}^{N \times N}$ & The attention matrix in the $k^\text{th}$ encoder layer \\ 
        
        $\widehat{\mathbf{C}}^{(k)} \in \mathbb{R}^{N \times N}$ & The attention matrix in the $k^\text{th}$ decoder layer \\ 
        
        ${\mathbf{h}_i}^{(k)} \in \mathbb{R}^{d^{(k)}}$ & The representation of node $i$ generated by the $k^\text{th}$ encoder layer \\
        
        $\widehat{\mathbf{h}}_i^{(k)} \in \mathbb{R}^{d^{(k)}}$ & The representation of node $i$ reconstructed by the $k^\text{th}$ decoder layer \\
        
        $\mathbf{h}_i \in \mathbb{R}^{d^{(L)}}$ & The representation of node $i$ ($\mathbf{h}_i = {\mathbf{h}_i}^{(L)} = \widehat{\mathbf{h}}_i^{(L)}$)   \\
        
        $\mathbf{x}_i \in \mathbb{R}^{F}$ & The features of node $i$ ($  \mathbf{h}_i^{(0)} = \mathbf{x}_i$) \\
        
        $\widehat{\mathbf{x}}_i \in \mathbb{R}^{F}$ & The reconstructed features of node $i$ ($\widehat{\mathbf{x}}_i = \widehat{\mathbf{h}}_i^{(0)}$) \\
        
        $ \alpha_{ij}^{(k)}$ & The attention coefficient indicating the relative relevance of neighboring node $j$ to node $i$ in the $k^\text{th}$ encoder layer \\
        
        $ \widehat{\alpha}_{ij}^{(k)}$ & The attention coefficient indicating the relative relevance of neighboring node $j$ to node $i$ in the $k^\text{th}$ decoder layer \\
        
        $\mathcal{N}_i$ & The neighborhood of node $i$, including itself \\
        
        \hline

    \end{tabular}
\end{table}

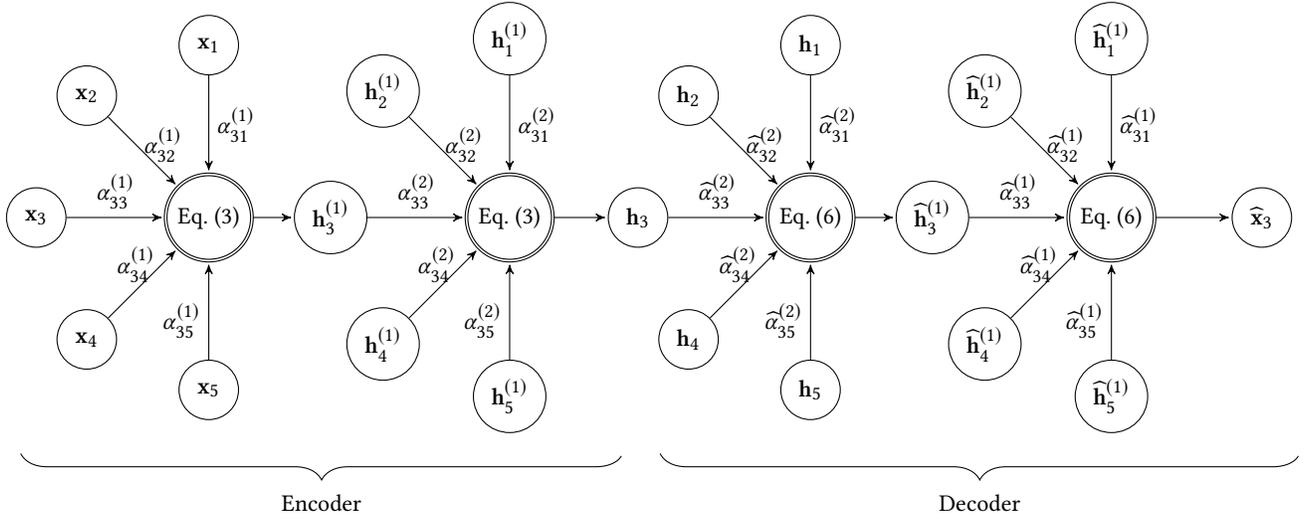
\begin{figure*}
    
    \begin{tikzpicture}[>=stealth',
    shorten > = 1pt,
    node distance = 1.3cm,
    every label/.append style = {font=\tiny}
    ]

    \draw [decorate,decoration={brace,amplitude=10pt,mirror,raise=4pt},yshift=0pt]
    (-2.5,-3) -- (5.5,-3) node [black,midway,yshift=-0.8cm] {Encoder};
    
    \draw [decorate,decoration={brace,amplitude=10pt,mirror,raise=4pt},yshift=0pt]
    (6,-3) -- (14.5,-3) node [black,midway,yshift=-0.8cm] {Decoder};
    
    \node (s1)  at (0,0) [state,accepting]    {Eq. (\ref{eq:encoder})};
    \node (x1) [state,above=of s1]    {$\mathbf{x}_1$};
    \node (x2) [state,above left=of s1]    {$\mathbf{x}_2$};
    \node (x3) [state,left=of s1]    {$\mathbf{x}_3$};
    \node (x4) [state,below left=of s1]    {$\mathbf{x}_4$};
    \node (x5) [state,below=of s1]    {$\mathbf{x}_5$};

    \node (s2)  at (4,0) [state,accepting]    {Eq. (\ref{eq:encoder})};
    \node (h11) [state,above=of s2]    {$\mathbf{h}_1^{(1)}$};
    \node (h21) [state,above left=of s2]    {$\mathbf{h}_2^{(1)}$};
    \node (h31) [state,left=of s2]    {$\mathbf{h}_3^{(1)}$};
    \node (h41) [state,below left=of s2]    {$\mathbf{h}_4^{(1)}$};
    \node (h51) [state,below=of s2]    {$\mathbf{h}_5^{(1)}$};

    \node (s3)  at (8,0) [state,accepting]    {Eq. (\ref{eq:decoder})};
    \node (h12) [state,above=of s3]    {$\mathbf{h}_1$};
    \node (h22) [state,above left=of s3]    {$\mathbf{h}_2$};
    \node (h32) [state,left=of s3]    {$\mathbf{h}_3$};
    \node (h42) [state,below left=of s3]    {$\mathbf{h}_4$};
    \node (h52) [state,below=of s3]    {$\mathbf{h}_5$};

    \node (s4)  at (12,0) [state,accepting]    {Eq. (\ref{eq:decoder})};
    \node (h-11) [state,above=of s4]    {$\widehat{\mathbf{h}}_1^{(1)}$};
    \node (h-21) [state,above left=of s4]    {$\widehat{\mathbf{h}}_2^{(1)}$};
    \node (h-31) [state,left=of s4]    {$\widehat{\mathbf{h}}_3^{(1)}$};
    \node (h-41) [state,below left=of s4]    {$\widehat{\mathbf{h}}_4^{(1)}$};
    \node (h-51) [state,below=of s4]    {$\widehat{\mathbf{h}}_5^{(1)}$};

    \node (x-3)  at (14,0) [state]    {$\widehat{\mathbf{x}}_3$};

    \path[->]
    
    (x1)  edge node[right, pos=0.5]    {$\alpha_{31}^{(1)}$}    (s1)
    (x2)  edge node[right, pos=0.4]    {$\alpha_{32}^{(1)}$}    (s1)
    (x3)  edge node[above, pos=0.5]    {$\alpha_{33}^{(1)}$}    (s1)
    (x4)  edge node[above, pos=0.4]    {$\alpha_{34}^{(1)}$}    (s1)
    (x5)  edge node[left, pos=0.4]    {$\alpha_{35}^{(1)}$}    (s1)

    (s1)  edge node[left, pos=0.4]    {} (h31)
    
    (h11)  edge node[right, pos=0.5]    {$\alpha_{31}^{(2)}$}    (s2)
    (h21)  edge node[right, pos=0.4]    {$\alpha_{32}^{(2)}$}    (s2)
    (h31)  edge node[above, pos=0.5]    {$\alpha_{33}^{(2)}$}    (s2)
    (h41)  edge node[above, pos=0.4]    {$\alpha_{34}^{(2)}$}    (s2)
    (h51)  edge node[left, pos=0.4]        {$\alpha_{35}^{(2)}$}    (s2)
    
    (s2)  edge node[left, pos=0.4]    {} (h32)
    
    (h12)  edge node[right, pos=0.5]    {$\widehat{\alpha}_{31}^{(2)}$}    (s3)
    (h22)  edge node[right, pos=0.4]    {$\widehat{\alpha}_{32}^{(2)}$}    (s3)
    (h32)  edge node[above, pos=0.5]    {$\widehat{\alpha}_{33}^{(2)}$}    (s3)
    (h42)  edge node[above, pos=0.4]    {$\widehat{\alpha}_{34}^{(2)}$}    (s3)
    (h52)  edge node[left, pos=0.4]        {$\widehat{\alpha}_{35}^{(2)}$}    (s3)

    (s3)  edge node[left, pos=0.4]    {} (h-31)
    
    (h-11)  edge node[right, pos=0.5]    {$\widehat{\alpha}_{31}^{(1)}$}    (s4)
    (h-21)  edge node[right, pos=0.4]    {$\widehat{\alpha}_{32}^{(1)}$}    (s4)
    (h-31)  edge node[above, pos=0.5]    {$\widehat{\alpha}_{33}^{(1)}$}    (s4)
    (h-41)  edge node[above, pos=0.4]    {$\widehat{\alpha}_{34}^{(1)}$}    (s4)
    (h-51)  edge node[left, pos=0.4]    {$\widehat{\alpha}_{35}^{(1)}$}    (s4)
    
    (s4)  edge node[left, pos=0.4]    {} (x-3)
    
    ;
    \end{tikzpicture}
    \caption{ The illustration of reconstructing the features of node 3, with neighborhood $\mathcal{N}_3=\{1,2,3,4,5\}$, using the graph attention auto-encoder with 2 layers; we note that $\mathbf{h}_i^{(0)} = \mathbf{x}_i$, $\mathbf{h}_i = \mathbf{h}_i^{(2)} = \widehat{\mathbf{h}}_i^{(2)}$, and $\widehat{\mathbf{x}}_i = \widehat{\mathbf{h}}_i^{(0)}$ , $\forall i \in \{ 1,2,..,N \}$ .}
    \label{fig:architecture}
\end{figure*}

\subsection{Encoder}
The encoder in our architecture takes node features and generates node representations by using the graph structure through stacked layers. We use multiple encoder layers for two reasons. First, more layers make our model deeper, and hence increasing the learning capability. Second, they propagate node representations through the graph structure, resulting in richer node embeddings. 


Each encoder layer generates new representations of nodes by utilizing their neighbors' representations according to their relevance. To determine the relevance between nodes and their neighbors, we use a self-attention mechanism with shared parameters among nodes, following the work of Velickovic et al. \cite{velickovic2018graph}. In the $k^{\text{th}}$ encoder layer, the relevance of a neighboring node $j$ to node $i$ is computed as follows:

\begin{equation}
\label{eq:e_ij}
e_{ij}^{(k)} = \text{Sigmoid} \left({\mathbf{v}_s^{(k)}}^T \sigma \left( \mathbf{W}^{(k)} \mathbf{h}_i^{(k-1)} \right) + {\mathbf{v}_r^{(k)}}^T \sigma \left( \mathbf{W}^{(k)} \mathbf{h}_j^{(k-1)} \right) \right)
\end{equation}

\noindent where $\mathbf{W}^{(k)} \in \mathbb{R}^{d^{(k)} \times d^{(k-1)}}$, $\mathbf{v}_s^{(k)} \in \mathbb{R}^{d^{(k)}}$, and $\mathbf{v}_r^{(k)} \in \mathbb{R}^{d^{(k)}}$ are the trainable parameters of the $k^{\text{th}}$ encoder layer, $\sigma$ denotes the activation function and $\text{Sigmoid}$ represents the sigmoid function (i.e., $\text{Sigmoid }(x) = 1 / (1 + \text{exp}^{-x})$).

To make the relevance coefficients of node $i$'s neighbors comparable, we normalize them by using the softmax function as follows:

\begin{equation}
\label{eq:alpha_ij}
\alpha_{ij}^{(k)} = \frac{ \text{exp} \left( e_{ij}^{(k)} \right) }{ \sum_{l \in \mathcal{N}_i} \text{exp} \left( e_{il}^{(k)} \right) }
\end{equation}

\noindent where $\mathcal{N}_i$ represents the neighborhood of node $i$ (i.e., a set of nodes connected to node $i$ according to the adjacency matrix $\mathbf{A}$, including node $i$ itself).

By considering node features as initial node representations (i.e., ${\mathbf{h}_i}^{(0)} = \mathbf{x}_i$, $\forall i \in \{ 1,2,..,N \}$), the $k^\text{th}$ encoder layer generates the representation of node $i$ in layer $k$ as follows:

\begin{equation}
\label{eq:encoder}
\mathbf{h}_i^{(k)} = \sum_{j \in \mathcal{N}_i}  \alpha_{ij}^{(k)}  \sigma \left(\mathbf{W}^{(k)} \mathbf{h}_j^{(k-1)} \right)
\end{equation}
After applying $L$ encoder layers, we consider the output of the last layer as the final node representations (i.e., $\mathbf{h}_i = \mathbf{h}_i^{(L)}$, $\forall i \in \{ 1,2,..,N \}$).

\subsection{Decoder}
Our encoder is reminiscent of graph attention networks \cite{velickovic2018graph}, which use supervised learning to embed nodes. Our main contribution is reversing the encoding process in order to learn node representations without any supervision. To this end, we use a decoder with the same number of layers as the encoder. Each decoder layer attempts to reverse the process of its corresponding encoder layer. In other words, each decoder layer reconstructs the representations of nodes by utilizing the representations of their neighbors according to their relevance. The normalized relevance (i.e., attention coefficient) of a neighboring node $j$ to node $i$ in the  $k^{\text{th}}$ decoder layer is computed as follows:

\begin{equation}
\label{eq:decoder_alpha_ij}
\widehat{\alpha}_{ij}^{(k)} = \frac{ \text{exp} \left( \widehat{e}_{ij}^{(k)} \right) }{ \sum_{l \in \mathcal{N}_i} \text{exp} \left( \widehat{e}_{il}^{(k)} \right) }
\end{equation}

\begin{equation}
\label{eq:decoder_e_ij}
\widehat{e}_{ij}^{(k)} = \text{Sigmoid} \left( {{}\widehat{\mathbf{v}}_s^{(k)}}^T \sigma \left( \widehat{\mathbf{W}}^{(k)} \widehat{\mathbf{h}}_i^{(k)} \right) + {{}\widehat{\mathbf{v}}_r^{(k)}}^T \sigma \left( \widehat{\mathbf{W}}^{(k)} \widehat{\mathbf{h}}_j^{(k)} \right) \right)
\end{equation}

\noindent where $\widehat{\mathbf{W}}^{(k)} \in \mathbb{R}^{ d^{(k-1)} \times d^{(k)}}$, $\widehat{\mathbf{v}}_s^{(k)} \in \mathbb{R}^{d^{(k-1)}}$, and $\widehat{\mathbf{v}}_r^{(k)} \in \mathbb{R}^{d^{(k-1)}}$ are the trainable parameters of the $k^\text{th}$ decoder layer.

By considering the output of the encoder as the input of the decoder (i.e., $\widehat{\mathbf{h}}_i^{(L)} = \mathbf{h}_i^{(L)}$, $\forall i \in \{ 1,2,..,N \}$), the $k^\text{th}$ decoder layer reconstructs the representation of node $i$ in layer $k-1$ as follows:

\begin{equation}
\label{eq:decoder}
\widehat{\mathbf{h}}_i^{(k-1)} = \sum_{j \in \mathcal{N}_i}  \widehat{\alpha}_{ij}^{(k)}  \sigma \left( \widehat{\mathbf{W}}^{(k)} \widehat{\mathbf{h}}_j^{(k)}  \right)
\end{equation}


After applying $L$ decoder layers, we consider the output of the last layer as the reconstructed node features (i.e., $ \widehat{\mathbf{x}_i}= \widehat{\mathbf{h}}_i^{(0)}$, $\forall i \in \{ 1,2,..,N \}$). Figure \ref{fig:architecture} illustrates the process of reconstructing node features in GATE through an example.

\subsection{Loss Function}

Graph-structured data include node features and the graph structure, and both should be encoded by high-quality node representations. 
We minimize the reconstruction loss of node features as follows:

\begin{equation}
\label{eq:semantic_loss}
\sum_{i=1}^N{|| \mathbf{x}_i - \widehat{\mathbf{x}}_i||}_2
\end{equation}

The absence of an edge between two nodes in the graph does not necessarily imply dissimilarity due to the possibility of feature similarity. Thus, we minimize the reconstruction loss of the graph structure by making the representations of neighboring nodes similar. We accomplish this by minimizing the following equation:


\begin{equation}
\label{eq:structure_loss}
- \sum_{i=1}^N \sum_{j \in \mathcal{N}_i} \log \left( \frac{1}{1+ \text{exp}({-\mathbf{h}_i^T {\mathbf{h}_j})}} \right)
\end{equation}


By merging Eq. (\ref{eq:semantic_loss}) and Eq. (\ref{eq:structure_loss}), we minimize the reconstruction loss of node features and the graph structure as follows:

\begin{equation}
\text{Loss} = \sum_{i=1}^N{|| \mathbf{x}_i - \widehat{\mathbf{x}}_i||_2} - \lambda \sum_{j \in \mathcal{N}_i}\log \left( \frac{1}{1+ \text{exp}({-\mathbf{h}_i^T {\mathbf{h}_j})}} \right)
\end{equation}
\noindent where $\lambda$ controls the contribution of the graph structure reconstruction loss.

\subsection{Matrix Formulation}
Since the adjacency matrix $\mathbf{A}$  is usually very sparse in practice, we can leverage sparse matrix operations (e.g., sparse softmax) to tackle large graphs. Therefore, we present the corresponding matrix formulas for the aforementioned encoder and decoder equations.

Let us begin with obtaining the attention matrix $\mathbf{C}^{(k)} \in \mathbb{R}^{N \times N}$ in the $k^{\text{th}}$ encoder layer, where $\mathbf{C}_{ij}^{(k)} = \alpha_{ij}^{(k)}$ if there is an edge between node $i$ and node $j$, and $\mathbf{C}_{ij}^{(k)}=0$ otherwise. We compute $\mathbf{C}^{(k)}$as follows:

\begin{equation}
\label{eq:attention_matrix}
\mathbf{C}^{(k)} = \text{Softmax} \left( \text{Sigmoid} \left( \mathbf{M}_s^{(k)} +  \mathbf{M}_r^{(k)} \right)  \right)
\end{equation}


\begin{equation}
\mathbf{M}_s^{(k)} = \mathbf{A} \odot \left( {\mathbf{v}_s^{(k)}}^T \sigma \left( \mathbf{W}^{(k)} \mathbf{H}^{(k-1)} \right) \right)
\end{equation}
\begin{equation}
\mathbf{M}_r^{(k)} = \mathbf{A} \odot \left( {\mathbf{v}_r^{(k)}}^T \sigma \left( \mathbf{W}^{(k)} \mathbf{H}^{(k-1)} \right) \right)^T
\end{equation}

\noindent where $\odot$ is element-wise multiplication with broadcasting capability and $\sigma$ denotes the activation function. 

By considering $\mathbf{H}^{(0)} = \mathbf{X} $, the $k^\text{th}$ encoder layer generates node representations in layer $k$ as follows:
\begin{equation}
\mathbf{H}^{(k)} = \sigma \left( \mathbf{W}^{(k)} \mathbf{H}^{(k-1)} \right)  \mathbf{C}^{(k)}
\end{equation}

After applying $L$ encoder layers, we consider $\mathbf{H}^{(L)}$ as the final node representation matrix (i.e., $\mathbf{H} = \mathbf{H}^{(L)}$).

The attention matrix $\widehat{\mathbf{C}}^{(k)} \in \mathbb{R}^{N \times N}$ in the $k^{\text{th}}$ decoder layer, where $\widehat{\mathbf{C}}_{ij}^{(k)} = \widehat{\alpha}_{ij}^{(k)}$ if there is an edge between node $i$ and node $j$, and $\widehat{\mathbf{C}}_{ij}^{(k)}=0$ otherwise, is computed as follows:

\begin{equation}
\label{eq:decoder_attention_matrix}
\widehat{\mathbf{C}}^{(k)} = \text{Softmax} \left( \text{Sigmoid} \left( \widehat{\mathbf{M}}_s^{(k)} +  \widehat{\mathbf{M}}_r^{(k)} \right)  \right)
\end{equation}

\begin{equation}
\widehat{\mathbf{M}}_s^{(k)} = \mathbf{A} \odot \left( {{}\widehat{\mathbf{v}}_s^{(k)}}^T \sigma \left( \widehat{\mathbf{W}}^{(k)} \widehat{\mathbf{H}}^{(k)} \right) \right)
\end{equation}
\begin{equation}
\widehat{\mathbf{M}}_r^{(k)} = \mathbf{A} \odot \left( {{}\widehat{\mathbf{v}}_r^{(k)}}^T \sigma \left( \widehat{\mathbf{W}}^{(k)} \widehat{\mathbf{H}}^{(k)} \right) \right)^T
\end{equation}

By considering $\widehat{\mathbf{H}}^{(L)} = {\mathbf{H}}^{(L)} $, the $k^\text{th}$ decoder layer reconstructs node representations in layer $k-1$ as follows:
\begin{equation}
\widehat{\mathbf{H}}^{(k-1)} =  \sigma \left( \widehat{\mathbf{W}}^{(k)} \widehat{\mathbf{H}}^{(k)} \right)  \widehat{\mathbf{C}}^{(k)}
\end{equation}

After applying $L$ decoder layers, we consider $\widehat{\mathbf{H}}^{(0)}$ as the reconstructed node feature matrix (i.e., $ \widehat{\mathbf{X}}= \widehat{\mathbf{H}}^{(0)} $).

%

Algorithm \ref{alg:1} shows the forward propagation of our proposed architecture using matrix formulation. 


\begin{algorithm}[t]
    \caption{GATE forward propagation algorithm using matrix formulation.}\label{alg:1}
    \begin{algorithmic}[1]
        \Statex{\textbf{Input:} The node feature matrix $\mathbf{X}$ and the adjacency matrix $\mathbf{A}$}
        \Statex{\textbf{output:} The node representation matrix $\mathbf{H}$} and the reconstructed node feature matrix $\widehat{\mathbf{X}}$
        \State Initialize $\mathbf{W}^{(k)}$, $\widehat{\mathbf{W}}^{(k)}$, $\mathbf{v}_s^{(k)}$, $\widehat{\mathbf{v}}_s^{(k)}$, $\mathbf{v}_r^{(k)}$ and $\widehat{\mathbf{v}}_r^{(k)}$, $\forall k \in \{ 1,2,..,L \}$ 
        \State $\mathbf{H}^{(0)} = \mathbf{X}$
        
        \For{$epoch\gets 1 \textbf{\space to } P$}
        
        \For{$k\gets 1 \textbf{\space to } L$}
        \State  Compute $\mathbf{C}^{(k)}$ according to Eq. (\ref{eq:attention_matrix})
        \State $\mathbf{H}^{(k)} = \sigma \left( \mathbf{W}^{(k)} \mathbf{H}^{(k-1)} \right)  \mathbf{C}^{(k)}$
        \EndFor
        
        \State $\widehat{\mathbf{H}}^{(L)} = \mathbf{H}^{(L)}$
        
        \For{$k\gets L \textbf{\space to } 1$}
        \State  Compute $\widehat{\mathbf{C}}^{(k)}$ according to Eq. (\ref{eq:decoder_attention_matrix})
        \State $\widehat{\mathbf{H}}^{(k-1)} =  \sigma \left( \widehat{\mathbf{W}}^{(k)} \widehat{\mathbf{H}}^{(k)} \right)  \widehat{\mathbf{C}}^{(k)}$
        \EndFor

        \State $\mathbf{H} = \mathbf{H}^{(L)}$
        \State $\widehat{\mathbf{X}} = \widehat{\mathbf{H}}^{(0)}$
        
        
        \EndFor
        
    \end{algorithmic}
\end{algorithm}

\subsection{Complexity}
Our proposed auto-encoder is highly efficient because the operations involved in the graph attention mechanisms can be parallelized across edges, and the rest of the operations in the encoder and decoder can be parallelized across nodes. Theoretically, the time complexity of our architecture for one iteration can be expressed as follows:

\begin{equation}
O (N F D + E D)
\end{equation}
\noindent  where $N$ and $E$ are respectively the number of nodes and edges in the graph, $F$ is the number of node features and $D$ is the maximum $d^{(k)}$ in all layers (i.e., $D = \max_{ k \in \{1,2, .., L\}} d^{(k)}$). 

By taking advantage of sparse matrix operations, the space complexity of our auto-encoder is linear in terms of the number of nodes and edges.



\section{Evaluation}
\label{sec:evaluation}
In this section, we quantitatively and qualitatively evaluate the proposed GATE architecture using several benchmark datasets. Section \ref{datasets}, \ref{baselines}, and \ref{setup} respectively describe the datasets, baselines, and experimental setup used in our experiments. In Section \ref{results}, we quantitatively evaluate the efficacy of our architecture. Section \ref{in_depth} investigates the impact of the three main components used in our proposed architecture, namely the self-attention mechanism, graph structure reconstruction, and node feature reconstruction. Finally, we investigate the quality of the node representations learned by GATE in Section \ref{qualitaive}.

\subsection{Datasets}
\label{datasets}

For transductive tasks, we use three benchmark datasets---Cora, Citeseer and Pubmed \cite{sen2008collective}---that are widely used to evaluate attributed graph embedding methods.
 In all datasets, each node belongs to one class. We follow the experimental setup of Yang et al. \cite{yang2016revisiting}, where 20 nodes per class are used for training. In the transductive setup, we have access to the graph structure and all nodes' feature vectors during training. We evaluate the predictive performance of each method on 1000 test nodes; 500 additional nodes are also used for validation of supervised methods. 
The statistics of the datasets are presented in Table \ref{tab:datasets}. 

For inductive tasks, we also use the same datasets and experimental setup in order to evaluate the generalization power of different methods to unseen nodes by comparing the difference between their performance in transductive and inductive tasks for the same dataset. As required by inductive learning, any information related to (unseen) test nodes, including features and edges, are completely unobserved during training.

\subsection{Baselines}
\label{baselines}

We compare our proposed auto-encoder against the following state-of-the-art supervised and unsupervised methods:
\begin{itemize}
    \item  \textbf{DeepWalk} \cite{perozzi2014deepwalk}: DeepWalk is a graph embedding method, which trains Skipgram model \cite{mikolov2013distributed} on simulated random walks over the graph. 
    \item  \textbf{Enhanced DeepWalk (DeepWalk + features)}: This baseline is a variant of DeepWalk concatenating raw node features and DeepWalk embeddings to take advantage of node features.
    \item \textbf{Label Propagation (LP)} \cite{zhu2002learning}: LP assigns labels to unlabeled nodes by propagating available labels in the graph. 
    \item \textbf{Planetoid} \cite{yang2016revisiting}: This baseline is an attributed graph embedding, which learns node representations by predicting available class labels and the neighborhood context in the graph using random walks.
    \item \textbf{Chebyshev} \cite{defferrard2016convolutional}: This baseline utilizes graph convolutional networks with high-order fast localized convolutions.
    \item \textbf{Graph Convolutional Networks (GCN)} \cite{kipf2017semi}: GCN incorporates spectral convolutions into neural networks to learn node representations.
    \item \textbf{Graph Auto-Encoder (GAE)} \cite{kipf2016variational}: GAE uses graph convolutional networks as the encoder and reconstruct the graph structure in the encoder.
    \item \textbf{Variational Graph Auto-Encoder (VGAE)} \cite{kipf2016variational}: VGAE is the variational version of GAE.
    \item \textbf{GraphSAGE} \cite{hamilton2017inductive}:  GraphSAGE has four unsupervised variants, which differ in their feature aggregator as follows: GraphSAGE-GCN (applying a convolution-style aggregator), GraphSAGE-mean (taking the element-wise mean of feature vectors), GraphSAGE-LSTM (aggregating by providing neighboring nodes' features into a LSTM), and GraphSAGE-pool (performing an element-wise max-pooling operation after applying a fully-connected neural network).

    \item \textbf{Monet} \cite{monti2017geometric}: This baseline generalizes convolutional neural networks to graph-structured data.
    
    \item \textbf{Deep Graph Infomax (DGI)} \cite{velivckovic2018deep}: DGI is an unsupervised attributed graph embedding, which simultaneously estimates and maximizes the mutual information between the graph-structured input and learned high-level graph summaries. 
    \item \textbf{StoGCN} \cite{chen2018stochastic}: StoCGN is a control variate-based
    stochastic algorithm for graph convolutional networks, which uses neighborhood sampling and historical hidden representations to reduce the receptive field of the graph convolution.
    \item \textbf{Graph Attention Networks (GAT)} \cite{velickovic2018graph}: GAT utilizes an attention mechanism to determine the influence of neighboring nodes in final node representations.

\end{itemize}

For transductive tasks, we compare our auto-encoder against supervised and unsupervised approaches. Unsupervised baselines include DeepWalk, enhanced DeepWalk, VGAE, GAE, and DGI. Supervised approaches are LP, Planetoid, Chebyshev, Monet, CGN, StoGCN, and GAT. For inductive tasks, we similarly compare GATE against supervised and unsupervised approaches. Unsupervised baselines include VGAE, GAE, and four unsupervised variants of GraphSAGE. Supervised approaches are Planetoid and GAT.

\begin{table}
    \centering
    \caption{The statistics of the benchmark datasets.}
    \label{tab:datasets}
    \resizebox{\columnwidth}{!}{%
        \begin{tabular}{ l c c c c c c c } \hline
            \textbf{Dataset} &  \textbf{Nodes} & \textbf{Edges} & \textbf{Features} & \textbf{Classes} & \textbf{Train/Val/Test Nodes }  \\
            \hline \hline
            \textbf{Cora}  & 2,708 & 5,429 & 1,433 & 7 & 140/500/1,000 \\
            \textbf{Citeseer} & 3,327 & 4,732 & 3,703 & 6 & 120/500/1,000 \\    
            \textbf{Pubmed}  & 19,717 & 44,338 & 500 & 3 & 60/500/1,000 \\
            \hline
        \end{tabular}
    }
\end{table}

\begin{table*}
    \centering
    \caption{Node classification accuracies on the Cora, Citeseer and Pubmed datasets. The first column shows the type of data used during training for each method. Data types are the node feature matrix $\mathbf{X}$, adjacency matrix $\mathbf{A}$, and labels $\mathbf{Y}$.}
    \label{tab:accuracies}

    \begin{subtable}[t]{\textwidth}
        \centering
        \caption{Transductive}
        \label{tab:transductive_classification}
    \begin{tabular}{ l l c c c } \toprule
        \textbf{Available Data} & \textbf{Method} &  \textbf{Cora} & \textbf{Citeseer} & \textbf{Pubmed}   \\
        \hline
        \hline
        $\mathbf{X}$ & Raw features \hspace{3.58cm} & $47.9\pm0.4 \% $ & $49.4\pm0.2 \% $ & $69.1\pm0.3 \% $ \\
        $\mathbf{A}$ & DeepWalk (Perozzi et al. \cite{perozzi2014deepwalk}) & $67.2\%$ & $43.2\%$ & $65.3\%$ \\
        $\mathbf{A}$, $\mathbf{Y}$ & LP (Zhu et al. \cite{zhu2002learning}) & $68.0\%$ & $45.3\%$ & $63.0\%$ \\
        \hline
        $\mathbf{X}$, $\mathbf{A}$ &  DeepWalk + features  & $70.7 \pm0.6 \%$ & $51.4 \pm0.5 \%$ & $74.3 \pm0.9 \%$ \\
        $\mathbf{X}$, $\mathbf{A}$ &  VGAE (Kipf \& Welling \cite{kipf2016variational}) & $72.4 \pm0.2 \%$ & $55.7 \pm0.2 \%$ & $71.6 \pm0.4$ \% \\
        $\mathbf{X}$, $\mathbf{A}$ &  GAE (Kipf \& Welling \cite{kipf2016variational}) & $81.8 \pm0.1 \%$ & $69.2 \pm0.9 \%$ & $78.2 \pm0.1 \%$  \\
        $\mathbf{X}$, $\mathbf{A}$ & DGI (Velickovic et al. \cite{velivckovic2018deep}) & $82.3 \pm0.6 \% $ & $71.8 \pm0.7 \% $ & $76.8 \pm0.6 \% $ \\
        $\mathbf{X}$, $\mathbf{A}$ & \textbf{GATE} (ours) & $\mathbf{83.2} \pm0.6 \% $ & $71.8 \pm0.8 \% $ & $\mathbf{80.9} \pm0.3 \% $ \\
        \hline
        $\mathbf{X}$, $\mathbf{A}$, $\mathbf{Y}$ & Planetoid (Yang et al. \cite{yang2016revisiting}) & $75.7\%$ & $62.9\%$ & $75.7\%$ \\
        $\mathbf{X}$, $\mathbf{A}$, $\mathbf{Y}$ & Chebyshev (Defferrard et al. \cite{defferrard2016convolutional})  & $81.2\%$ & $69.8\%$ & $74.4\%$ \\
        $\mathbf{X}$, $\mathbf{A}$, $\mathbf{Y}$ & Monet (Monti et al. \cite{monti2017geometric}) & $81.7 \pm0.5 \% $ & --- & $78.0 \pm0.3 \% $ \\
        $\mathbf{X}$, $\mathbf{A}$, $\mathbf{Y}$ & GCN (Kipf \& Welling \cite{kipf2017semi}) & $81.5\%$ & $70.3\%$ & $79.0\%$ \\
        $\mathbf{X}$, $\mathbf{A}$, $\mathbf{Y}$ &  StoGCN (Chen et al. \cite{chen2018stochastic}) & $82.0 \pm0.8 \% $ & $70.9 \pm0.2 \% $ & $79.0 \pm0.4 \% $ \\
        $\mathbf{X}$, $\mathbf{A}$, $\mathbf{Y}$ & GAT (Velickovic et al. 
        \cite{velickovic2018graph}) & $83.0 \pm0.7 \% $ & $\mathbf{72.5} \pm0.7 \% $ & $79.0 \pm0.3 \% $ \\
        \bottomrule
    \end{tabular}
    \end{subtable}
    
    \hspace{\fill}
    \begin{subtable}[t]{\textwidth}
        
        \centering
        \caption{Inductive}
        \label{tab:inductive_classification}
        \begin{tabular}{ l l c c c } \toprule
            \textbf{Available Data} & \textbf{Method} &  \textbf{Cora} & \textbf{Citeseer} & \textbf{Pubmed}   \\
            \hline
            \hline
            
            $\mathbf{X}$, $\mathbf{A}$ & GraphSAGE-LSTM (Hamilton et al. \cite{hamilton2017inductive}) & $50.1 \pm0.2 \%$ & $40.3 \pm0.2 \%$ & $77.1 \pm0.1 \%$\\
            $\mathbf{X}$, $\mathbf{A}$ & GraphSAGE-pool (Hamilton et al. \cite{hamilton2017inductive})  & $57.5 \pm0.2 \%$ & $45.9 \pm0.2 \%$ & $79.9 \pm0.1 \%$ \\
            $\mathbf{X}$, $\mathbf{A}$ &  VGAE (Kipf \& Welling \cite{kipf2016variational}) & $58.4 \pm0.4 \%$ & $55.4 \pm0.2 \%$ & $71.1 \pm0.2 \%$ \\
            $\mathbf{X}$, $\mathbf{A}$ & GraphSAGE-mean (Hamilton et al. \cite{hamilton2017inductive}) & $67.0 \pm0.2 \%$ & $52.8 \pm0.1 \%$ & $79.3 \pm0.1 \%$\\
            $\mathbf{X}$, $\mathbf{A}$ & GraphSAGE-GCN (Hamilton et al. \cite{hamilton2017inductive}) & $74.3 \pm0.1 \%$ & $54.5 \pm0.1 \%$ & $77.5 \pm0.1 \%$\\
            $\mathbf{X}$, $\mathbf{A}$ &  GAE (Kipf \& Welling \cite{kipf2016variational}) & $80.5 \pm0.1 \%$ & $69.1 \pm0.9 \%$ &  $78.1 \pm0.2 \%$ \\
            $\mathbf{X}$, $\mathbf{A}$ & \textbf{GATE} (ours) & $\mathbf{82.5} \pm0.5 \% $ & $\mathbf{71.5} \pm0.7 \% $ & $\mathbf{80.8} \pm0.3 \% $ \\
            \hline
            $\mathbf{X}$, $\mathbf{A}$, $\mathbf{Y}$ & Planetoid (Yang et al. \cite{yang2016revisiting}) & $61.2\%$ & $64.7\%$ & $77.2\%$ \\
            $\mathbf{X}$, $\mathbf{A}$, $\mathbf{Y}$ & GAT (Velickovic et al. \cite{velickovic2018graph}) & $76.4 \pm0.2 \% $ & $66.4 \pm0.2 \% $ & $77.7 \pm0.03 \% $ \\
            
            \bottomrule
        \end{tabular}
    \end{subtable}

\end{table*}

\subsection{Experimental Setup}
\label{setup}
In our experiments, Adam optimizer \cite{kingma2014adam} is used to learn model parameters with an initial learning rate of $10^{-4}$. For all datasets, we use two layers with 512 node representation dimensions (i.e., $d^{(1)}=d^{(2)}=512$). We set the number of epochs to 100 for Cora and Citeseer, and 500 for Pubmed. We also set $\lambda$ to 0.5 for Cora and Pubmed, and 20 for Citeseer. We use only half of the trainable parameters by setting $\widehat{\mathbf{W}}^{(k)} = {\mathbf{W}^{(k)}}^T$ and $\widehat{\mathbf{C}}^{(k)} = \mathbf{C}^{(k)}$. Moreover, $\sigma$ is set to the identity function, empirically resulting in better performance compared to other activation functions. Tensorflow \cite{abadi2016tensorflow} is used to implement GATE. 


For the baselines to which we directly compare GATE, we use their default hyperparameter settings as well as the following settings. We perform a hyperparameter sweep on initial learning rates $ \{10^{-3}, 10^{-4}, 10^{-5}, 10^{-6}, 10^{-7}  \}$ and $\{ 10^{-2}, 10^{-3}, 10^{-4} \}$ for unsupervised and supervised methods  respectively. We also swept over the number epochs in the set $\{ 50, 100, 200, 300 \}$ for VGAE and GAE due to their sensitivity to this hyperparameter. For supervised methods, we perform a sweep on dropouts $\{ 0, 0.2, 0.5 \}$. We also set the number of node representation dimensions to 512 for all of these baselines.

\begin{figure*}[!htb]
	
	\centering
	\begin{subfigure}[b]{0.49\linewidth}
		\begin{tikzpicture}
		\begin{axis}[
		ybar,
		enlargelimits=0.15,
		legend style={at={(0.5,-0.15)},
			anchor=north,legend columns=-1},
		xtick={1,...,3},
		ytick={60,65,...,85},
		ymin=65,
		ymax=85,
		xticklabels={Cora,Pubmed,Citeseer},
		nodes near coords align={vertical},
		]
		\addplot +[
		postaction={
			pattern=crosshatch,
			pattern color=black!70,
		},
		draw=none,
		point meta=y,
		every node near coord/.style={inner ysep=5pt},
		error bars/.cd,
		y dir=both,
		y explicit]
		table [y error=error] {
			x   y           error    label
			1   83.2   0.6   1 
			2   80.9   0.3   2
			3   71.8   0.8   3 
		};

		\addplot +[
		postaction={
			pattern=dots,
			pattern color=black!70,
		},
		draw=none,
		point meta=y,
		every node near coord/.style={inner ysep=5pt},
		error bars/.cd,
		y dir=both,
		y explicit]
		table [y error=error] {
			x   y           error    label
			1   82.1   0.8   1 
			2   77.4   0.9   2
			3   69.5   0.9   3 
		};
		
		\addplot +[green!50!black,fill=green!70!black!50,
		postaction={
			pattern=north west lines,
			pattern color=black!70,
		},
		draw=none,
		point meta=y,
		every node near coord/.style={inner ysep=5pt},
		error bars/.cd,
		y dir=both,
		y explicit]
		table [y error=error] {
			x   y         error    label
			1   81.6   0.5   1 
			2   74.9   0.4   2
			3   71.4   0.7   3 
		};

		\addplot +[black!60!black,fill=black!60!black!30,
		postaction={
			pattern=north east lines,
			pattern color=black!70,
		},
		draw=none,
		point meta=y,
		every node near coord/.style={inner ysep=5pt},
		error bars/.cd,
		y dir=both,
		y explicit]
		table [y error=error] {
			x   y           error    label
			1   76.8   0.8   1
			2   74.4   0.9   2
			3   67.5   0.6   3
		};
		
		\legend{GATE,GATE/F,GATE/S,GATE/A}
		\end{axis}
		\end{tikzpicture}
		\subcaption{Trandsuctive}
		\label{fig:in_depth_transductive}
	\end{subfigure}
	\begin{subfigure}[b]{0.49\linewidth}
		\begin{tikzpicture}
		\begin{axis}[
		ybar,
		enlargelimits=0.15,
		legend style={at={(0.5,-0.15)},
			anchor=north,legend columns=-1},
		xtick={1,...,3},
		ytick={60,65,...,85},
		ymin=65,
		ymax=85,
		xticklabels={Cora,Pubmed,Citeseer},
		nodes near coords align={vertical},
		]
		\addplot +[
		postaction={
			pattern=crosshatch,
			pattern color=black!70,
		},
		draw=none,
		point meta=y,
		every node near coord/.style={inner ysep=5pt},
		error bars/.cd,
		y dir=both,
		y explicit]
		table [y error=error] {
			x   y           error    label
			1   82.5   0.5   1 
			2   80.8   0.3   2
			3   71.5   0.7   3 
		};

		\addplot +[
		postaction={
			pattern=dots,
			pattern color=black!70,
		},
		draw=none,
		point meta=y,
		every node near coord/.style={inner ysep=5pt},
		error bars/.cd,
		y dir=both,
		y explicit]
		table [y error=error] {
			x   y           error    label
			1   81.7   0.8   1 
			2   74.6   0.2   2
			3   68.8   0.99   3 
		};
	
		\addplot +[green!50!black,fill=green!70!black!50,
		postaction={
			pattern=north west lines,
			pattern color=black!70,
		},
		draw=none,
		point meta=y,
		every node near coord/.style={inner ysep=5pt},
		error bars/.cd,
		y dir=both,
		y explicit]
		table [y error=error] {
			x   y         error    label
			1   81.3   0.5   1 
			2   74.8   0.4   2
			3   71.4   0.6   3 
		};
		
		\addplot +[black!60!black,fill=black!60!black!30,
		postaction={
			pattern=north east lines,
			pattern color=black!70,
		},
		draw=none,
		point meta=y,
		every node near coord/.style={inner ysep=5pt},
		error bars/.cd,
		y dir=both,
		y explicit]
		table [y error=error] {
			x   y           error    label
			1   75.8   0.9   1 
			2   74.2   0.9   2
			3   66.9   0.7   3
		};
		
		\legend{GATE,GATE/F,GATE/S,GATE/A}
		\end{axis}
		\end{tikzpicture}
		\subcaption{Inductive}
		\label{fig:in_depth_inductive}
	\end{subfigure}
	
	\caption{Node classification accuracies  on the Cora, Citeseer and Pubmed datasets for four variants of our architecture.}
	\label{fig:in_depth}
\end{figure*}
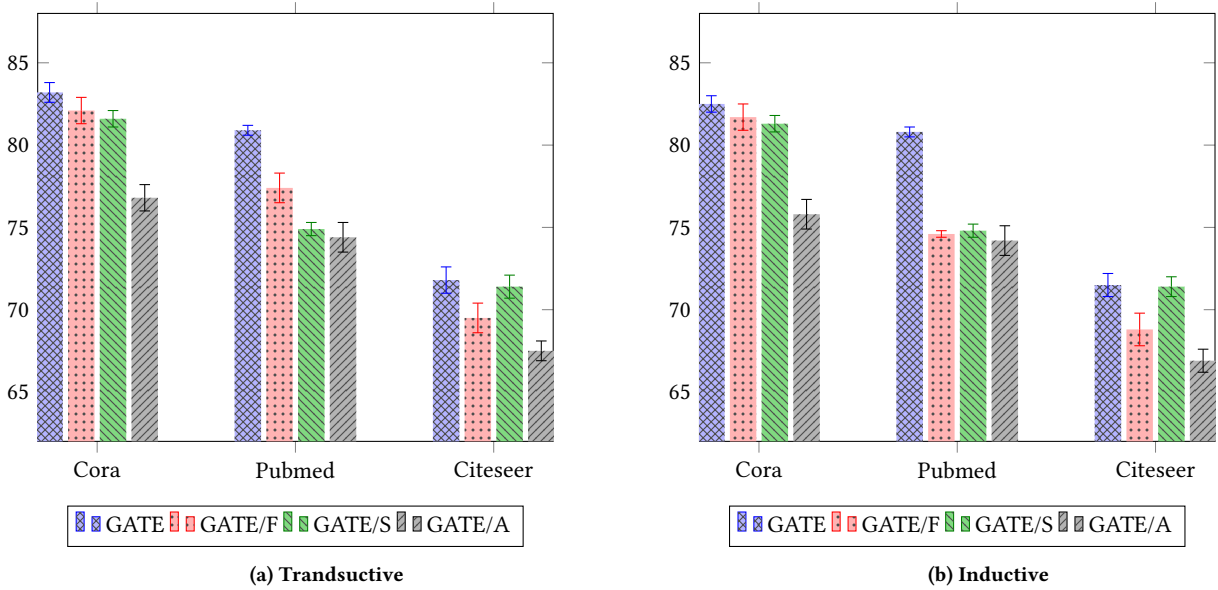

\subsection{Comparison}
\label{results}
In this section, we compare out proposed method with the aforementioned state-of-the-art baselines based on transductive and inductive node classifications. For transductive node classification, we report the mean classification accuracy (with standard deviation) of our method on the test nodes after 100 runs of training (followed by logistic regression). The accuracies for GCN, DeepWalk, and LP are retrieved from Kipf \& Welling \cite{kipf2017semi}. We also reuse the metrics reported in Velickovic et al. \cite{velivckovic2018deep} for the performance of enhanced DeepWalk, DGI, and logistic regression with raw features. The accuracies for GAT, Chebyshev, and Monet are taken from Velickovic et al. \cite{velickovic2018graph}. For StoGCN and Planetoid, the metrics are retrieved from their papers \cite{chen2018stochastic, yang2016revisiting}. Moreover, we directly compare our method against GAE and VGAE. 

Table \ref{tab:transductive_classification} shows the transductive node classification accuracies for the Cora, Citeseer, and Pubmed datasets. Accordingly, we make the following observations:

\begin{itemize}
    \item GATE achieves strong performance across all three datasets. Particularly, GATE outperforms all supervised and unsupervised baselines on the Cora and Pubmed datasets.
    \item Our unsupervised architecture is competitive with the performance of the best supervised baseline (i.e., GAT), even improving upon it by a margin of 1.9\% and 0.2\% on Pubmed and Cora respectively.
    \item GATE outperforms or matches all unsupervised baselines across all datasets. We observe an improvement of 2.7\% and 0.9\% over the best unsupervised baselines for Pubmed and Cora respectively.
    \item For the Citeseer dataset, the accuracy of GATE follows that of GAT. This can be attributed to the low average node degree of 1.4 for Citeseer---which is lower than Cora's (2) and Pubmed's (2.25). The scarcity of neighbors and the abundance of features can give the supervised baselines (i.e., GAT), which take advantage of a supervised loss, leverage over unsupervised methods 
    \item The reconstruction of node features by GATE results in a considerable improvement compared to the graph auto-encoder baselines reconstructing only the graph structure. Compared to the best graph auto-encoder baseline (i.e., GAE), we achieve an improvement gain of 2.7\%, 2.6\%, and 1.4\% on Pubmed, Citeseer, and Cora respectively.

\end{itemize}

For inductive node classification, we utilize the same datasets used for the transductive tasks. This enables us to compare the performance of GATE between transductive and inductive tasks for the same dataset in order to evaluate the generalization power of our auto-encoder to unseen nodes. We report the mean classification accuracy (with standard deviation) of our method on the (unseen) test nodes after 100 runs of training (followed by logistic regression). For Planetoid, its accuracies are retrieved from Yang et al. \cite{yang2016revisiting}. We directly compare our method against VGAE, GAE, GAT, and four variants of GraphSAGE.

Table \ref{tab:inductive_classification} shows the inductive node classification accuracies for the Cora, Citeseer, and Pubmed datasets. Accordingly, we make the following observations:

\begin{itemize}
    \item GATE exceeds the performance of all supervised and unsupervised baselines across all three datasets. We are able to improve upon the best baselines by a margin of 2.4\%, 2\%, and 0.9\% on Citeseer, Cora, and Pubmed respectively.
    \item We can observe that GATE achieves similar accuracies for inductive and transductive tasks with regard to the same dataset. For example, the accuracy difference between inductive and transductive tasks is 0.1\%, 0.3\%, and 0.7\%  on Pubmed, Citeseer, and Cora respectively. This suggests that GATE naturally generalizes to unseen nodes.
    \item Unlike GATE, not every method performing well on transductive tasks (e.g., GAT) can perform well on inductive tasks.
\end{itemize}


\subsection{In-depth Analysis}
\label{in_depth}
In this section, we investigate the impact of the three main components used in our proposed architecture, namely the self-attention mechanism, graph structure reconstruction and node feature reconstruction. In our experiments, we use the following variants of our architecture: 

\begin{itemize}
	\item \textbf{GATE}: The full version of our proposed auto-encoder which includes all three components.
	\item \textbf{GATE/A}: A variant of our architecture which includes all components except the self-attention mechanism. In other words, we assign the same importance to each neighbor.
	\item \textbf{GATE/S}: A variant of our architecture which includes all components except the graph structure reconstruction. 
	\item \textbf{GATE/F}: A variant of our architecture which includes all components except the node feature reconstruction.  
\end{itemize}

We first compare the four variants of our architecture based on transductive node classification. Figure \ref{fig:in_depth_transductive} shows the mean classification accuracy (with standard deviation) of all four variants on the test nodes after 100 runs of training (followed by logistic regression).  Accordingly, we make the following observations: 

\begin{itemize}
	\item GATE outperforms other variants in all datasets. Therefore, each component contributes to the overall performance of our architecture.  
	\item GATE/A performs worse than other variants. This suggests that the self-attention mechanism contributes the most in our architecture compared to the graph structure and node feature reconstructions.
	\item In Cora and Pubmed which have higher average node degree (i.e., 2 and 2.5 respectively), GATE/F outweighs the performance of GATE/S. On the other hand, GATE/S exceeds the performance of GATE/F in Citeseer which has the lowest average node degree (i.e., 1.4) and the highest number of features.
\end{itemize}

\begin{figure*}
	\centering
	\begin{subfigure}{.5\textwidth}
		\centering
		\includegraphics[width=.99\linewidth,trim=4 4 4 4,clip]{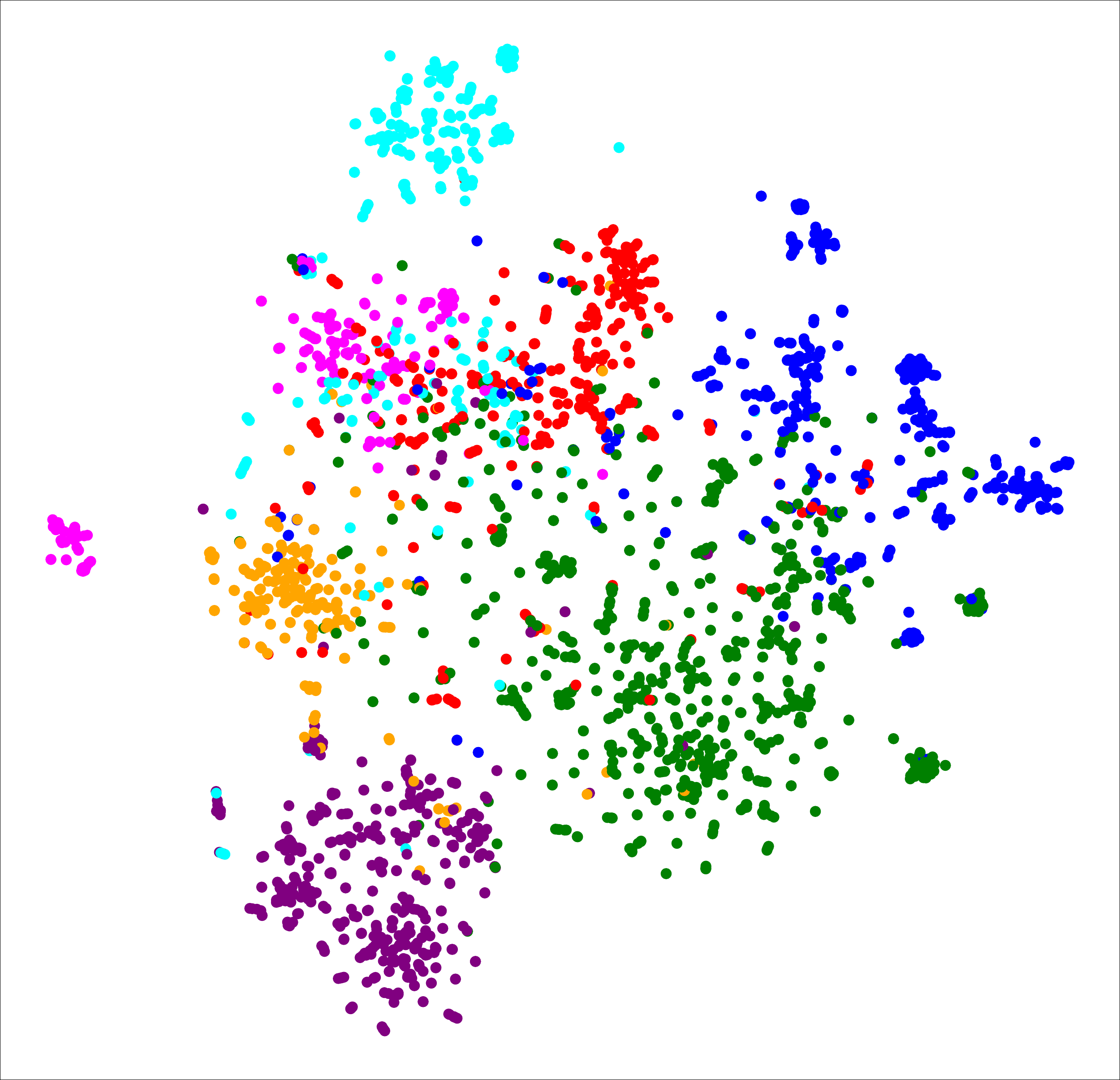}
		\caption{The t-SNE visualization of nodes.}
		\label{fig:visualization_1}
	\end{subfigure}%
	\begin{subfigure}{.5\textwidth}
		\centering
		\includegraphics[width=.99\linewidth,trim=4 4 4 4,clip]{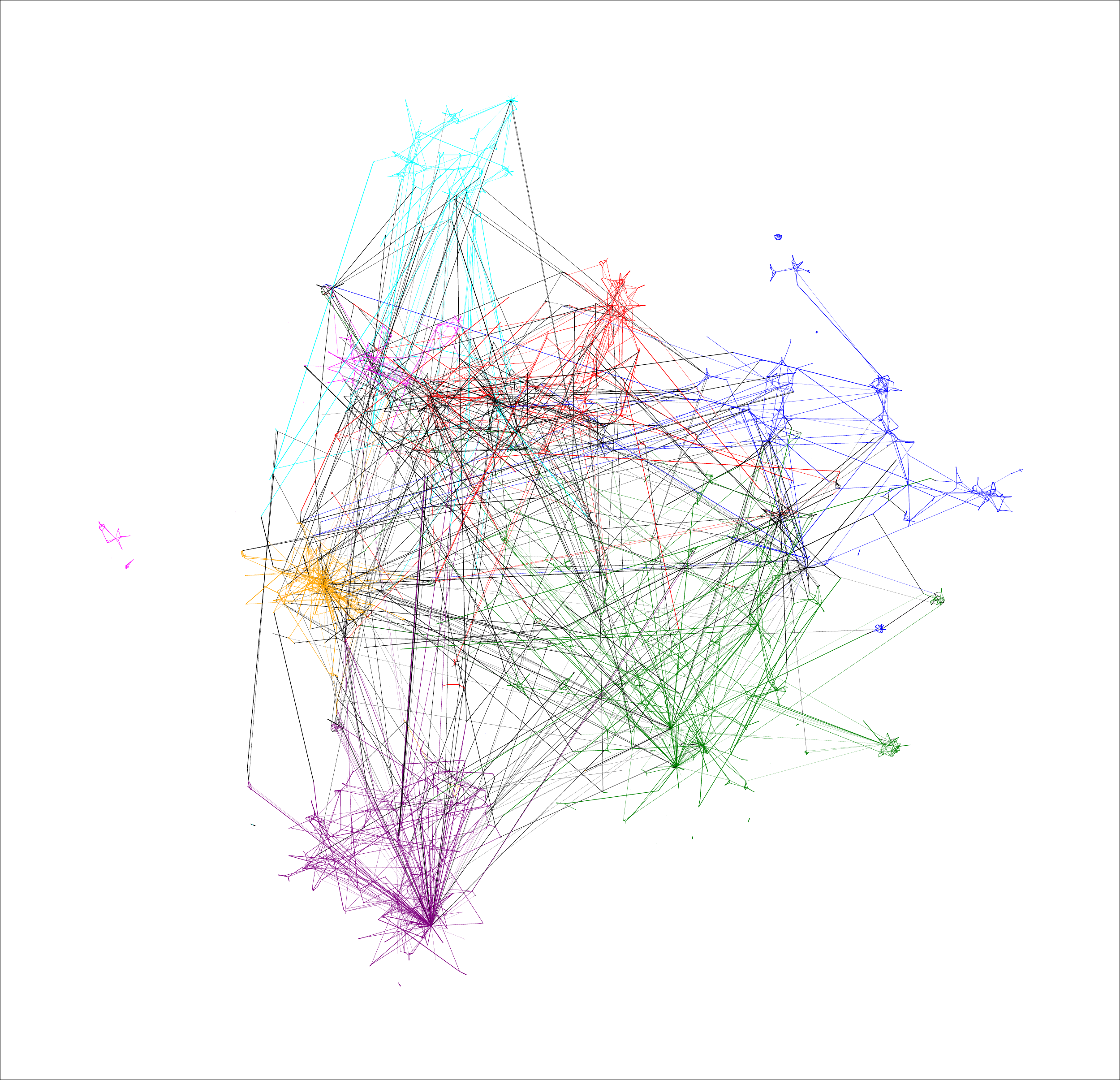}
		\caption{The t-SNE visualization of edges.}
		\label{fig:visualization_2}
	\end{subfigure}
	
	\caption{The t-SNE visualizations of the node representations learned by GATE on the Cora dataset in node and edge perspectives. In Figure \ref{fig:visualization_1}, node colors denote classes. In Figure \ref{fig:visualization_2}, the edges with source and target nodes belonging to the same class are colored with the corresponding color of the class, and the others are colored black. Moreover, edge thickness indicates the averaged attention coefficients between node $i$ and $j$ across all layers (i.e., $\sum_{k=1}^{L} \left( \alpha_{ij}^{(k)} + \widehat{\alpha}_{ij}^{(k)} \right) / 2L$ )}
	
	\label{fig:visualization}
\end{figure*}

Now we compare all variants of our architecture based on inductive node classification. Figure \ref{fig:in_depth_inductive} shows the mean classification accuracy (with standard deviation) of all four variants on the (unseen) test nodes after 100 runs of training (followed by logistic regression).  Accordingly, we make the following observations:

\begin{itemize}
	\item Like the transductive node classification experiments, GATE and GATE/A are respectively the best and the worst variants of our architecture in all datasets.
	\item We observe that the performances of GATE/F and GATE/S  in Cora and Citeseer are similar to those of transductive node classification experiments. However, we notice a huge drop in the performance of GATE/F in Pubmed even though the performance of GATE/S has not undergone such a decrease. This can be attributed to both the low number of features and high average node degree of Pubmed compared to those of Cora and Citeseer, which hugely benefit GATE/F in transductive learning over inductive learning. 
\end{itemize}



\subsection{Qualitative Analysis}
\label{qualitaive}
In this section, we qualitatively investigate the effectiveness of the node representations and attention coefficients learned by GATE. To this end, we utilize t-SNE \cite{maaten2008visualizing} to project the learned node representations into a two-dimensional space. Due to space limitation, we only show the visualization for the Cora dataset. Figure \ref{fig:visualization_1} shows the t-SNE visualization of the learned node representations for Cora, where node colors denote classes. We can observe that the learned node representations result in discernible clusters. 

Figure \ref{fig:visualization_2} shows the t-SNE visualization of the edges, in the Cora dataset, thickened by their attention coefficients averaged across all layers. In this figure, the edges with source and target nodes belonging to the same class are colored with the color of the class, and the others are colored black. Accordingly, we expect high-quality node representations to result in thicker colorful edges. In Figure \ref{fig:visualization_2}, we can observe that the colorful edges are usually thicker than the black edges. However, in few spots where GATE faces difficulty in separating nodes belonging to different classes, we can notice the presence of some thick black edges.

\section{Conclusion}
\label{sec:conclusions}
Graph-structured data can be found in many real-world scenarios, such as social media \cite{salehi2018sentiment, salehi2018detecting}, protein-protein interaction networks \cite{velickovic2018graph}, and citation networks \cite{kipf2017semi}.

In this paper, we have introduced the graph attention auto-encoder (GATE), a novel neural architecture for unsupervised representation learning on graph-structured data. 
By stacking multiple encoder/decoder layers equipped with graph attention mechanisms, GATE is the first graph auto-encoder, which reconstructs both node features and the graph structure.

Experiments on both transductive and inductive tasks using three benchmark datasets demonstrate the efficacy of GATE, which learns high-quality node representations. In most experiments, our auto-encoder outweighs state-of-the-art supervised and unsupervised baselines. Moreover, our experiments show that GATE naturally generalizes to unseen nodes.

The tensor manipulation framework (i.e., Tensorflow \cite{abadi2016tensorflow}) we used does not support matrix multiplication for rank-3 tensors, limiting batching capability of our auto-encoder. Therefore, addressing this limitation is an important direction for future work. 



\balance

\bibliographystyle{ACM-Reference-Format}
\bibliography{sample-sigconf}
\end{document}